\documentclass{article} % For LaTeX2e
\usepackage{iclr2027_conference,times}
\usepackage{amsmath,amsfonts,bm}

\def\eqref#1{equation~\ref{#1}}
\def\1{\bm{1}}

\DeclareMathAlphabet{\mathsfit}{\encodingdefault}{\sfdefault}{m}{sl}
\SetMathAlphabet{\mathsfit}{bold}{\encodingdefault}{\sfdefault}{bx}{n}

\usepackage[utf8]{inputenc} % allow utf-8 input
\usepackage[T1]{fontenc}    % use 8-bit T1 fonts
\usepackage{url}            % simple URL typesetting
\usepackage{booktabs}       % professional-quality tables
\usepackage{amsfonts}       % blackboard math symbols
\usepackage{nicefrac}       % compact symbols for 1/2, etc.
\usepackage{microtype}      % microtypography
\usepackage{xcolor}         % colors
\usepackage{amsmath}
\usepackage{subcaption}
\usepackage{multirow}
\usepackage{xcolor,colortbl,graphicx}
\definecolor{navyblue}{RGB}{0,0,128}
\definecolor{citecolor}{HTML}{0071BC}
\definecolor{linkcolor}{HTML}{ED1C24}
\definecolor{urlcolor}{RGB}{180, 100, 255}
\definecolor{oursrow}{RGB}{252, 230, 230}  % soft sage green
\usepackage{pifont}
\newcommand{\cmark}{\ding{51}}
\newcommand{\xmark}{\ding{55}}
\newsavebox{\tabboxA}
\newsavebox{\tabboxB}
\newsavebox{\tabboxC}
\usepackage{adjustbox}
\newlength{\tabrowht}
\usepackage[
    colorlinks=true,
    citecolor=citecolor,
    linkcolor=linkcolor,
    urlcolor=urlcolor,
]{hyperref}

\definecolor{ava}{RGB}{150, 0, 255}

\title{Render to Reason: Novel-View Semantic Prediction Improves Spatial Understanding in VLMs}

\author{%
  Yuqun Wu$^{1}$ \qquad
  Yao Xiao$^{1}$ \qquad
  Chuhang Zou$^{2}$ \qquad
  Shenlong Wang$^{1}$ \qquad
  Derek Hoiem$^{1}$ \\
  $^{1}$University of Illinois at Urbana-Champaign \qquad $^{2}$Meta
}

\iclrfinalcopy

\begin{document}

\maketitle
\fancyhead{}                          % remove "Published as a conference paper at ICLR 2027"
\renewcommand{\headrulewidth}{0pt}    % remove the horizontal line under the header

% \begin{abstract}
% Recent works augment Vision-Language Models (VLMs) with geometry features from pretrained 3D models, expecting that the geometric signal will boost spatial reasoning. However, we find that simply fusing geometry features and training on standard spatial QA yields only marginal improvements on high-level multi-hop tasks. We trace this to a shortcut behavior: visual features and language priors are already sufficient to answer most questions, so the model defaults to the visual pathway and underuses the geometric one. To address this, we propose \textbf{novel-view semantic rendering} as an auxiliary training task that requires the model to predict the semantic layout of an unobserved viewpoint, drawing inspiration from cognitive science: humans solve spatial reasoning tasks by mentally simulating novel viewpoints. This task is solvable only when both pathways are used together: geometry provides pose-dependent visibility, while vision provides semantic content. We evaluate on VSI-Bench, ReVSI, and a new low-level spatial QA benchmark that isolates geometric reasoning from object grounding. Our auxiliary task yields consistent improvements over the geometry-augmented baseline across all three benchmarks (+1.6 on VSI-Bench, +2.2 on ReVSI, +2.9 on low-level), and our full model achieves competitive performance with models trained on more data.
% \end{abstract}

\begin{figure}[h]
\centering
\vspace{-1.2em}
% \hspace{-3em}
\includegraphics[width=1.0\textwidth]{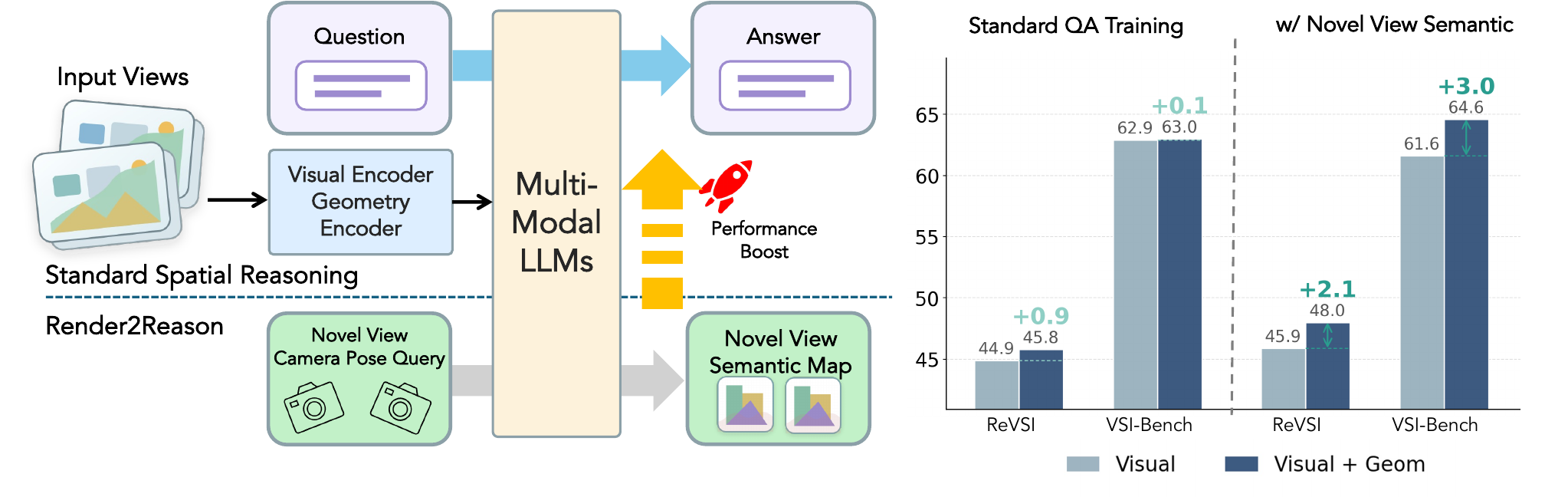}
\vspace{-2em}
\caption{
    % \textbf{Ambiguous Case.} When the second can is moved elsewhere and the third can is moved into the second can’s original location, our method only predicts that the third can has moved and cannot correctly infer the swap.
    \textbf{Render to Reason.} 
    % Left: We introduce an auxiliary training task, i.e., novel-view semantic rendering, which takes the input images and camera tokens from unseen views, and render the semantic layout of that view. Right: On ReVSI~\citep{zhang2026revsi} and VSI-Bench~\citep{Yang2024ThinkingIS}, adding geometry features under standard QA training yields marginal gains (+0.9, +0.1). Combined with our novel-view semantic rendering, the geometry feature benefits more (+2.1, +3.0), indicating that the auxiliary task encourages the model to integrate geometry features more effectively.
    Left: We introduce novel-view semantic rendering as an auxiliary training task, which takes input images and a target camera token, and renders the semantic layout of the unseen view. Right: On ReVSI~\citep{zhang2026revsi} and VSI-Bench~\citep{Yang2024ThinkingIS}, 
    % adding geometry features under standard QA training yields only marginal gains (+0.9, +0.1). Combined with our novel-view semantic rendering, geometry features contribute more (+2.1, +3.0), indicating that the auxiliary task encourages the model to integrate the geometry features more effectively.
    adding geometry features under standard QA training yields only marginal gains (+0.9, +0.1). When trained with our novel-view semantic rendering, adding geometry features yields larger gains (+2.1, +3.0), showing that the auxiliary task encourages the model to integrate geometry features more effectively.
}
\label{fig:teaser_figure}
% \vspace{-0.1em}
\end{figure}

\begin{abstract}
% \vspace{-0.3em}
Recent works augment Vision-Language Models with geometry features from pretrained 3D models, expecting that the geometric signal will boost spatial reasoning. However, we find that simply fusing geometry features and training on standard spatial QA yields only marginal improvements on high-level multi-hop tasks. We attribute this gap to a training-signal problem: standard spatial QA can be largely answered from visual features and language priors, so the geometry pathway receives weak gradients and fails to integrate with the visual features. To provide a training signal that requires geometry, we propose \textbf{novel-view semantic rendering} as an auxiliary training task that requires the model to predict the semantic layout of an unobserved viewpoint, inspired by humans' ability to mentally simulate novel viewpoints during spatial reasoning. This task encourages joint use of both pathways: geometry provides pose-dependent visibility, while vision provides semantic content. 
Our auxiliary task yields consistent improvements over the geometry-augmented baseline across all three benchmarks (up to +1.6 on VSI-Bench, +2.2 on ReVSI, +2.9 on our 3D-Point-QA dataset) 
and our full model surpasses prior open-source methods on VSI-Bench and on ReVSI.
Project page: 
\href{https://yuqunw.github.io/Render2Reason/}{https://yuqunw.github.io/Render2Reason/}.
\end{abstract}
\vspace{-0.3em}
\section{Introduction}
\vspace{-0.3em}
% Spatial reasoning from visual observations is important for embodied AI systems, robotic navigation, and augmented reality applications. Given a set of images of a scene, the system needs to answer spatial reasoning questions, e.g., estimating distance and direction between objects, counting items, and planning routes. Vision-Language Models (VLMs)~\citep{openai2024gpt4o, Comanici2025Gemini2P, wang2025internvl3, Qwen3-VL} have emerged as a promising architecture for such tasks due to their strong generalizability.
% To improve VLMs' spatial reasoning abilities, recent works~\citep{wu2025spatialmllmboostingmllmcapabilities, Fan2025VLM3RVM, Zheng2025LearningFV, Zhao2025SpaceMindCM, Zhang2026SSRPT} have augmented standard visual features with geometry features encoded from pretrained models~\citep{Wang2025VGGTVG, wang2025continuous} that support dense geometry predictions such as point clouds and camera poses, expecting that the geometric signal would boost spatial reasoning.

Spatial reasoning from visual observations is important for embodied AI, robotic navigation, and augmented reality applications, and Vision-Language Models (VLMs)~\citep{openai2024gpt4o, Comanici2025Gemini2P, wang2025internvl3, Qwen3-VL} have emerged as a promising architecture for such tasks due to their strong generalizability. To improve VLMs' spatial reasoning abilities, recent works~\citep{wu2025spatialmllmboostingmllmcapabilities, Fan2025VLM3RVM, Zheng2025LearningFV, Zhao2025SpaceMindCM, Zhang2026SSRPT} have augmented standard visual features with geometry features encoded from pretrained geometry models~\citep{Wang2025VGGTVG, wang2025continuous} that support dense geometry predictions such as point clouds and camera poses, expecting that the geometric signal would boost spatial reasoning. However, this expectation does not hold in practice: simply fusing geometry features and training on standard spatial QA yields only marginal improvements over a vision-only baseline on high-level multi-hop tasks (Fig.~\ref{fig:teaser_figure}).

% However, we find that this intuition does not straightforwardly hold. Simply fusing geometry features and training on standard spatial QA yields only marginal improvements over a vision-only baseline on high-level tasks (Fig.~\ref{fig:teaser_figure}).
We attribute this gap to a training-signal problem: standard spatial QA can be largely answered from visual features and language priors, so the geometry pathway is underused and fails to integrate with the visual features. 
To encourage stronger use of geometry, we propose \textbf{novel-view semantic rendering} as an auxiliary training task that requires both pathways. 
Our design draws inspiration from classic findings in cognitive science: humans solve spatial reasoning tasks by \textit{mentally simulating} novel viewpoints~\citep{Shepard1971MentalRO,Hegarty2004MechanicalRB}. We adapt this perspective to VLMs: if the model is trained to predict the semantic layout of an unseen viewpoint, it must combine the geometric pathway (\textit{which surfaces are visible from the target pose}) with the visual pathway (\textit{what those surfaces are}), and neither alone is sufficient (Fig.~\ref{fig:semantic_ablation}). 
The task can be supervised at scale from any video, with semantic labels generated by a pretrained semantic segmentation model~\citep{wang2023internimage}, and requires no additional human annotation.

% We evaluate our approach on existing multi-hop spatial reasoning datasets VSI-Bench~\citep{Yang2024ThinkingIS} and ReVSI~\citep{zhang2026revsi}, and the 3D-Point-QA dataset we introduce that isolates geometric reasoning from object grounding. Our experiments show that geometry features alone yield clear gains on low-level tasks (+1.7 points) but only marginal improvement on high-level multi-hop reasoning (+0.9 on ReVSI). Adding novel-view semantic rendering as auxiliary supervision improves both task types, with +1.6 on VSI-Bench, +2.2 on ReVSI, and +2.9 on the low-level benchmark. Our method achieves 67.3 average on VSI-Bench and 51.8 on ReVSI, matching models of similar scale while using less annotated training data.

We evaluate on the multi-hop spatial reasoning benchmarks VSI-Bench~\citep{Yang2024ThinkingIS} and ReVSI~\citep{zhang2026revsi}, and on \textbf{3D-Point-QA}, a low-level benchmark we introduce that isolates geometric reasoning from object grounding via point prompts. Adding novel-view semantic rendering as auxiliary supervision improves a geometry-augmented baseline by +1.6 on VSI-Bench, +2.2 on ReVSI, and +2.9 on 3D-Point-QA. Our full model achieves 75.3 average on VSI-Bench and 54.5 on ReVSI, surpassing prior open-source methods on ReVSI and on VSI-Bench.

In summary, our contributions are:
\begin{itemize}
    \item We propose \textbf{novel-view semantic rendering}, an auxiliary task that requires both visual and geometric reasoning during training and consistently improves spatial QA performance over a geometry-augmented baseline.
    \item Our full model outperforms prior open-source methods on VSI-Bench and ReVSI.
    \item We introduce \textbf{3D-Point-QA}, a low-level spatial reasoning dataset that isolates geometric reasoning from object grounding through point-prompt queries.
\end{itemize}
% In summary, our contributions are:
% \begin{itemize}
%     \item We identify a \textbf{training-signal problem} in geometry-augmented VLMs and propose \textbf{novel-view semantic rendering}, an auxiliary task that requires both visual and geometric reasoning during training. It consistently improves spatial QA performance over a geometry-augmented baseline and can be supervised from any video without additional human annotation.
%     \item We introduce \textbf{3D-Point-QA}, a low-level spatial reasoning dataset that isolates geometric reasoning from object grounding through point-prompt queries.
%     \item Our full model outperforms prior open-source methods of comparable scale on ReVSI and matches the strongest baselines on VSI-Bench, while using less annotated training data.
% \end{itemize}
\section{Related Work}

\subsection{Vision-Language Models for Spatial Understanding}
\label{sec:realted_3dvlm}
Building on general-purpose Vision-Language Models (VLMs)~\citep{Li2024LLaVAOneVisionEV, wang2025internvl3, Qwen3-VL}, recent works have specialized these architectures for spatial reasoning by integrating explicit 3D representations into VLMs. 3D-LLM~\citep{hong20233d} feeds 3D point cloud features into the language model, enabling 3D scene captioning, grounding, and question answering. Follow-up work explores point cloud encoders~\citep{xu2024pointllm, Chen2023LL3DAVI} and scene graph inputs~\citep{zhu20233d}. These methods assume access to dense 3D inputs at inference, motivating a line of work that targets the same spatial competencies from images alone.

To remove this dependency, recent methods~\citep{wu2025spatialmllmboostingmllmcapabilities, Fan2025VLM3RVM,Zheng2025LearningFV,Zhang2026SSRPT,Zhao2025SpaceMindCM,li2026thinking} inject geometry features extracted by pretrained 3D models~\citep{Wang2025VGGTVG, wang2024dust3r, wang2025continuous} directly into VLMs from images alone. These methods report consistent gains on spatial benchmarks, demonstrating that geometric features carry useful signal. ~\citet{chen2026think} and \citet{huang20263drs} improve spatial reasoning by training the VLM to predict geometry features at input views, distilling knowledge without taking geometry as direct input. In this work we further examine how this signal is integrated during training, and observe that under standard QA supervision the visual pathway dominates while the geometry pathway remains underused on high-level multi-hop questions. \citet{zhang2026make} are motivated by a similar concern and encourage greater use of geometry features through random appearance feature dropping and geometry-guided fusion. We instead target the training signal itself with an auxiliary objective that requires both pathways.
Closer to our work, several recent methods add predictive auxiliary objectives to VLM training for spatial reasoning. 
On the generation side~\citep{cao2024motion2vecsets, cao2026freeorbit4d,tang2026motion2vecsets}, 
% Ross3D~\citep{wang2025ross3d} reconstructs masked views and bird's-eye-view images, and 
Omni-View~\citep{hu2025omni} jointly trains 3D scene generation and understanding. Cambrian-S~\citep{Yang2025CambrianSTS} trains a next-latent-frame predictor to drive long-video memory, and Cambrian-P~\citep{yang2026cambrianpposegroundedvideounderstanding} additionally supervises camera pose as an auxiliary output. 
Our approach, given the camera pose of an arbitrary unseen view, instead predicts the discrete semantic layout via patch-level classification rather than generating appearance or predicting pose, supervised jointly with the QA objective.

\subsection{Spatial Reasoning Datasets and Benchmarks}\label{sec:realted_3dvlm_dataset}
Early 3D vision-language datasets focus on scene captioning and question answering with ground-truth 3D inputs, including ScanQA~\citep{azuma2022scanqa}, SQA3D~\citep{ma2022sqa3d}, and 3D-LLM's instruction data~\citep{hong20233d}. More recent benchmarks evaluate spatial reasoning directly from videos or image sequences without requiring 3D scans: VSI-Bench~\citep{Yang2024ThinkingIS} introduces eight categories of multi-hop spatial questions, ViewSpatial-Bench~\citep{li2025viewspatial} and MindCube~\citep{wang2025spatial} target perspective-taking and mental rotation, and OmniSpatial~\citep{Jia2025OmniSpatial} broadens coverage with cognitive-psychology categories such as spatial logic and dynamic reasoning. On the training side, several large-scale datasets have been released to improve spatial VLMs: SPAR~\citep{Zhang2025FromFT} provides 7M spatial QA pairs, VLM-3R~\citep{Fan2025VLM3RVM} extracts 208K reconstruction-aware QAs, and VSI-590K~\citep{Yang2025CambrianSTS} contributes 590K samples spanning multiple datasets~\citep{Dai2017ScanNetR3, Yeshwanth2023ScanNetAH, Dehghan2021ARKitScenesAD}. ReVSI~\citep{zhang2026revsi} refines existing 3D annotations and provides ground truth that depends on the input frame count, enabling more controlled evaluation. These datasets cover a wide range of multi-hop questions to evaluate high-level spatial reasoning. We complement these benchmarks with 3D-Point-QA, a low-level benchmark that extends DepthLM's~\citep{cai2025depthlm} single-image point-prompt design to multi-image queries (including cross-image point matching and 3D coordinate mapping), isolating geometric reasoning from text-based grounding.
\section{Method}
\begin{figure}[t]
\centering
\vspace{-.5em}
\includegraphics[width=1.0\textwidth]{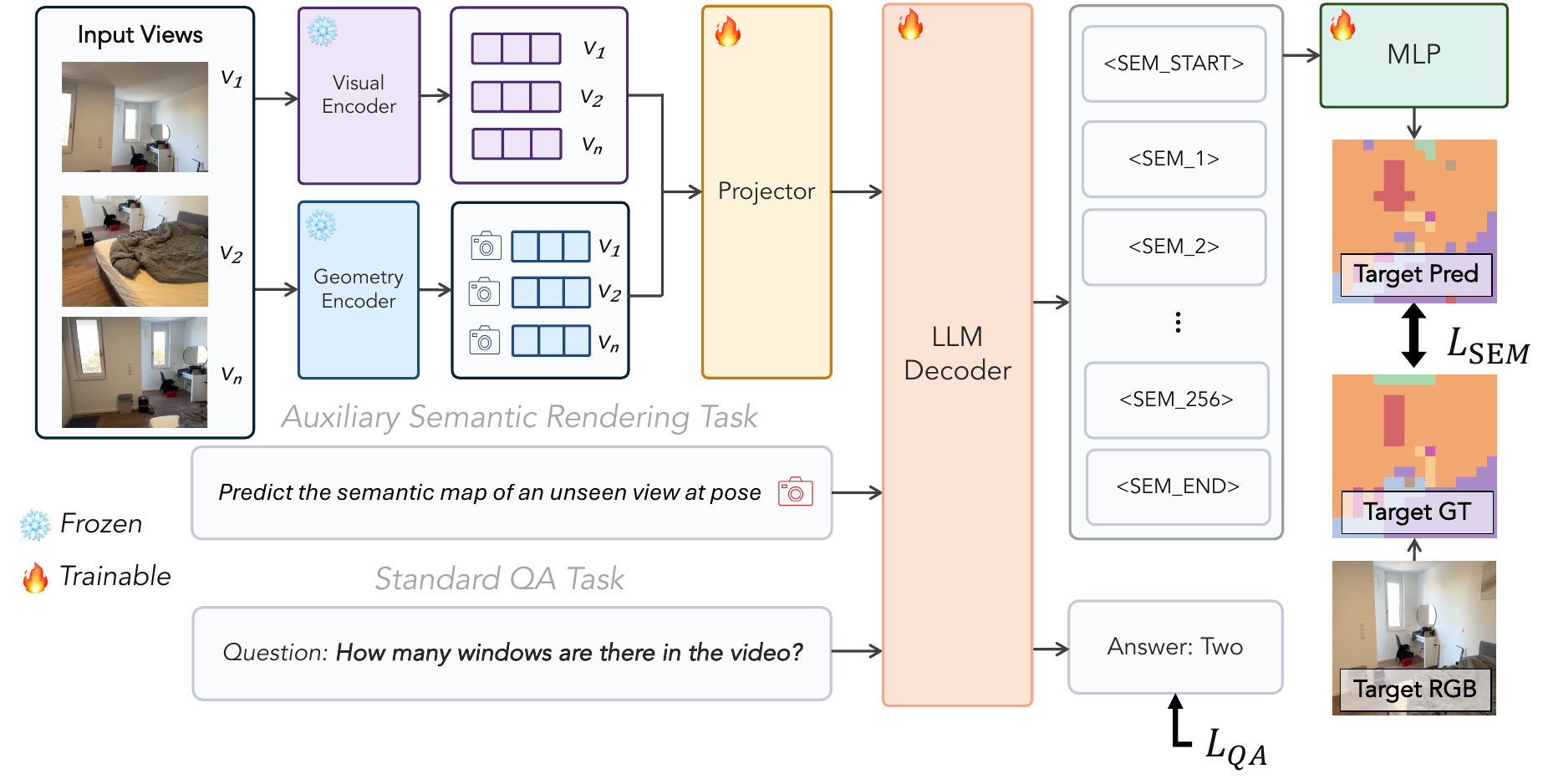}
\vspace{-2.em}
\caption{
    % \textbf{Ambiguous Case.} When the second can is moved elsewhere and the third can is moved into the second can’s original location, our method only predicts that the third can has moved and cannot correctly infer the swap.
    \textbf{Model Architecture.} Our model jointly supports semantic novel-view rendering and standard QA, sharing a backbone and differing only in prompt and output head. Input views are encoded into visual and geometry features, fused via cross-attention, and passed to the LLM decoder. For novel view semantic rendering, we extract a camera token from the target view via the geometry encoder and prepend it to the prompt; the LLM produces 256 output tokens, one per target patch in raster order, each mapped to a semantic class by a 2-layer MLP. For standard QA, the prompt only contains the question, and the LLM generates the answer without using the rendering MLP.
    % \yao{could you demonstrate the fusion module is cross-attention in the figure? The current version seems like concat}
}
\label{fig:architecture}
\vspace{-1.em}
\end{figure}

Our method (Fig.~\ref{fig:architecture}) targets spatial reasoning given multiview images. We first describe the base architecture with geometry features(\S\ref{sec:preliminaries}), then present our novel-view semantic rendering task (\S\ref{sec:semantic_rendering}), and the overall training procedure (\S\ref{sec:training}).

\subsection{Preliminaries}
\label{sec:preliminaries}

\paragraph{VLMs with geometry features.} Standard VLMs take $N$ input views $\{I_i\}_{i=1}^{N}$ and extract per-view visual tokens $\mathbf{v}_i \in \mathbb{R}^{L_v \times d_v}$ through a vision encoder (e.g., SigLIP~\citep{Zhai2023SigmoidLF, Tschannen2025SigLIP2M}). These tokens are passed through a learned projector $f_{\text{proj}}$ that maps them into the language model's hidden dimension, and then concatenated with text tokens $\mathbf{t}_{\text{text}}$ representing the question or instruction. The decoder autoregressively generates the answer given the concatenated tokens.

While standard VLMs exhibit strong general visual understanding, the visual encoder is primarily trained for high-level semantic understanding, and is not well suited for geometry reasoning tasks.
Recent works (Sec.~\ref{sec:realted_3dvlm}) augment the VLMs with geometry features. A pretrained geometry encoder VGGT~\citep{Wang2025VGGTVG} jointly processes all $N$ views to produce per-view geometry patch tokens $\mathbf{g}_i \in \mathbb{R}^{L_g \times d_g}$ and camera tokens $\mathbf{c}_i \in \mathbb{R}^{1 \times d_g}$, encoding dense geometry and camera pose information respectively.
To fuse geometry information into the visual representations, we concatenate the geometry and camera tokens as keys and values in a cross-attention module, with the visual tokens as queries, followed by a gated MLP with a residual connection:
\begin{equation}
    \mathbf{v}_i' = \mathbf{v}_i + \alpha \cdot \texttt{MLP}\!\left(\texttt{CrossAttn}\!\left(Q=\mathbf{v}_i,\; K=[\mathbf{g}_i; \mathbf{c}_i],\; V=[\mathbf{g}_i; \mathbf{c}_i]\right)\right),
\end{equation}
where $[\cdot\,;\cdot]$ denotes concatenation along the token dimension and $\alpha$ is a learnable gate. The fused visual tokens are then projected and fed to the decoder, allowing its predictions to incorporate dense 3D structure and camera-pose information encoded in the geometry features.
% We compared this fusion strategy against direct addition and decoder-level concatenation, and found cross-attention perform the best while keeping the LLM input length unchanged.
We compared this fusion strategy against direct addition and decoder-level concatenation, and found that cross-attention performs best while keeping the LLM input length unchanged. We use a decoder-concatenation variant only for the modality ablation in Fig.~\ref{fig:semantic_ablation}, since it allows each pathway to be removed at inference by dropping the corresponding tokens.

% \paragraph{Limitations of standard QA training.} 
% Standard QA training, however, provides limited gradient signal for the geometry pathway. A model trained with decoder-concatenation variant achieves 61.4 average accuracy on VSI-Bench~\citep{Yang2024ThinkingIS}, but still scores 60.2 when geometry features are removed at inference, indicating that the model relies primarily on the visual pathway under QA-only training. Visual features already contain implicit spatial cues sufficient to answer many spatial questions, leaving the standard QA objective with weak signal through the geometry pathway.

\paragraph{Limitations of standard QA training.} Standard QA training, however, provides limited training signal for the geometry pathway. Adding VGGT features to a VLM trained on the same QA data improves VSI-Bench scores by only +0.1 and +1.0 and ReVSI by +0.9 and +0.8 on InternVL3.5-4B and Qwen3-VL-4B respectively (Tab.~\ref{tab:ablation}). Visual features already contain implicit spatial cues sufficient to answer many spatial questions, leaving the standard QA objective with weak signal through the geometry pathway.

\subsection{Novel-View Semantic Rendering}
\label{sec:semantic_rendering}
\begin{figure}[t]
\vspace{-1em}
\centering
\includegraphics[width=1.0\textwidth]{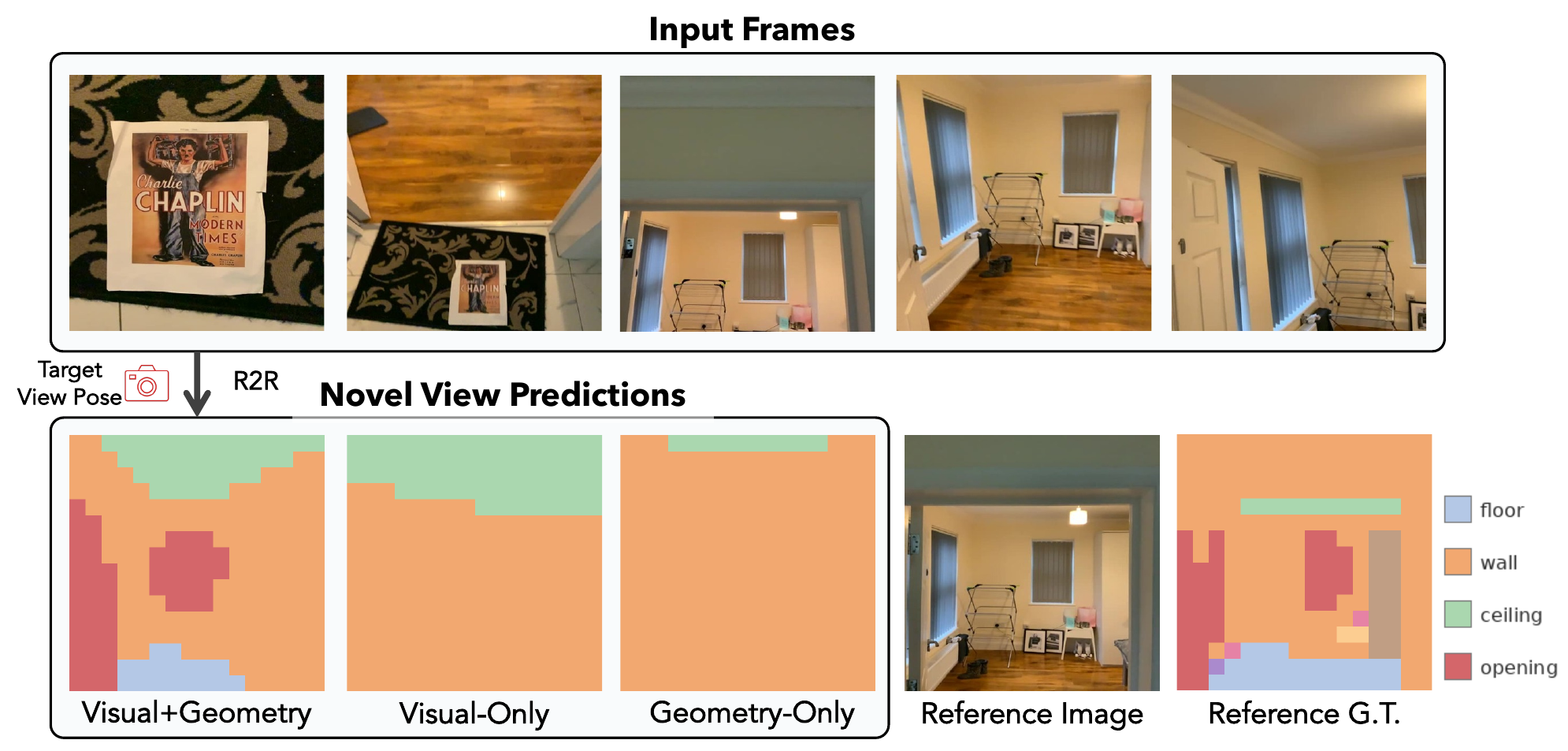}
\vspace{-1.5em}
\caption{
    % \textbf{Ambiguous Case.} When the second can is moved elsewhere and the third can is moved into the second can’s original location, our method only predicts that the third can has moved and cannot correctly infer the swap.
    \textbf{Modality Ablation for Novel View Semantic Rendering.} The model predicts a semantic map of an unseen viewpoint, given input frames and the target camera pose. Top row: input frames. Bottom row: target view. With both modalities, the prediction matches the target geometry. With vision only, the prediction fails to reach the target pose and \textbf{resembles the last input frame}, indicating that the geometry features are required to render at the correct viewpoint.
}
\label{fig:semantic_ablation}
\vspace{-1.em}
\end{figure}

\paragraph{Motivation.} To encourage the model to use both pathways, we design an objective that cannot be solved with either alone. Predicting the semantic layout from an unseen viewpoint satisfies this requirement: determining what is visible from the novel pose requires dense 3D geometry and camera pose reasoning, captured by the geometry features, and determining the semantic class of each visible surface requires the appearance and category information provided by the visual features.
To verify this, we train the decoder-concatenation variant with semantic rendering and remove each pathway while keeping the target camera token at inference. As shown in Fig.~\ref{fig:semantic_ablation}, visual-only predictions fail to localize the target view, while geometry-only predictions fail to produce reasonable class labels.
Simpler alternatives, such as better geometry--visual alignment or predicting VGGT features at novel views, do not improve spatial QA, since they do not require the model to combine both pathways (Appendix~\ref{app:motivation}).
We choose semantic layouts over RGB because semantic prediction couples scene structure with object identity, providing a signal aligned with spatial reasoning, while remaining a discrete prediction target that integrates naturally with the LLM's next-token prediction machinery.

% \yao{I feel this could be moved to ablation study section} 
% Before this design, we tested two alternatives. First, to test whether the weak geometry signal stems from misalignment between geometry and visual features, we pretrained the geometry projector with geometry-only captioning and instruction following before joint spatial QA training. This improved geometry-only inference but not joint inference, indicating that better alignment alone does not lead the model to combine the two pathways. Second, we predicted VGGT latent features at novel views, supervised through depth and poses decoded by the frozen VGGT heads. This reduced VSI-Bench accuracy by 1.5 points relative to the geometry-augmented baseline, likely because the target lies entirely in the geometry feature space and can be predicted from the geometry pathway alone. 

\paragraph{Task formulation.} We formulate the task as predicting the semantic map from the target viewpoint given the $N$ input views and a target camera token $\tilde{\mathbf{c}}_t$ corresponding to an unseen view. 
We obtain $\tilde{\mathbf{c}}_t$ by including the target frame as an additional input to the VGGT encoder alongside the $N$ input views, and extracting the target frame's camera token. 
We keep VGGT frozen and discard the target view's output patch tokens, retaining only its camera token and the context-view features for the auxiliary prediction task, and we verify that target-frame information leaking into the context features is negligible (Appendix~\ref{app:leakage}).
Since our goal is to encourage geometric reasoning rather than fine-grained rendering, we keep the output simple to classify once the viewpoint is correctly resolved.
By default, we use a 16$\times$16 grid (256 query tokens), which balances object merging at coarser resolutions against sequence length at finer ones (Tab.~\ref{tab:abl_grid}).
We pair this resolution with a compact taxonomy of $C=9$ room-level classes: \texttt{floor}, \texttt{wall}, \texttt{ceiling}, \texttt{opening} (doors, windows), \texttt{table}, \texttt{seating}, \texttt{cabinet}, \texttt{object}, and \texttt{void}. Each cell of the grid corresponds to the most common semantic class in that spatial region of the target view.

\paragraph{Model output and supervision.} To produce the semantic grid, a natural choice would be to extend the language model's vocabulary with $C$ class tokens and have the decoder autoregressively generate the $16 \times 16 = 256$ semantic labels. However, under teacher forcing, ground-truth labels of previously predicted patches are fed as input, making the task too easy: the model can infer each patch from neighboring labels, bypassing cross-modal reasoning. 
To prevent this shortcut and enable parallel decoding without overloading the LLM head with patch-level classification, we instead (i) add 256 learnable query tokens (\texttt{<SEM\_1>} to \texttt{<SEM\_256>}) to the vocabulary, one per spatial location in the 16$\times$16 grid, following common practice in referring segmentation~\citep{lai2024lisa, ren2024pixellm}; and (ii) attach a separate two-layer MLP head $f_{\text{sem}} : \mathbb{R}^{d} \to \mathbb{R}^{C}$ for class prediction. 
The query tokens are appended to the input sequence after the input views and the target camera token. The LLM decoder produces a hidden representation $\mathbf{h}_j \in \mathbb{R}^{d}$ for each query token $j$ in a single forward pass, and $f_{\text{sem}}$ maps each hidden representation to a class distribution in parallel:
$
    p_j = \mathrm{softmax}\!\left(f_{\text{sem}}(\mathbf{h}_j)\right), j = 1, \ldots, 256.
$
The task is supervised with standard cross-entropy loss:
\begin{equation}
    \mathcal{L}_{\text{sem}} = -\frac{1}{|\mathcal{V}|} \sum_{j \in \mathcal{V}} \log p_j[y_j],
\end{equation}
where $y_j$ is the ground-truth class at patch $j$, and $\mathcal{V}$ is the set of valid non-void patches.

\paragraph{Inference behavior.} The model's behavior at inference depends on which task is invoked. For spatial QA, the model takes the question as input and generates the answer autoregressively, leaving the semantic query tokens and MLP head unused. For novel-view rendering, the prompt contains the target camera token followed by the 256 query tokens. The LLM then produces hidden representations in a single forward pass, and $f_{\text{sem}}$ maps them to class labels for all patches in parallel.

\paragraph{Data generation.}
For each sample, we extract six views from consecutive video frames with random temporal intervals of 0.5 to 1.5 seconds. One view is randomly selected as the target, and the remaining five serve as inputs. We empirically find that this simple sampling strategy is more effective than explicitly maximizing viewpoint diversity for improving geometric reasoning performance, likely because it preserves realistic temporal coherence.
We generate semantic labels at scale using a pretrained segmentation model (Mask2Former~\citep{cheng2021mask2former} with the InternImage-H backbone~\citep{wang2023internimage}).
The predicted segmentation maps are downsampled to a $16 \times 16$ grid via max-pooling and mapped to the 9-class taxonomy by assigning the dominant class for each cell.

\subsection{Training}
\label{sec:training}

\paragraph{Joint objective.}
We jointly train the model on spatial QA and novel-view semantic rendering. During training, each batch contains a mixture of QA samples and rendering samples, so the model receives supervision from both tasks at every update. For each sample, we compute only the loss associated with its task: QA samples contribute to the QA loss, while rendering samples contribute to the semantic rendering loss. The overall objective is:
\begin{equation}    \mathcal{L} = \mathcal{L}_{\text{QA}} + \lambda_{\text{sem}} \mathcal{L}_{\text{sem}},
\end{equation}
where $\mathcal{L}_{\text{QA}}$ is the standard autoregressive token-level cross-entropy loss on the answer tokens for QA samples, $\mathcal{L}_{\text{sem}}$ is the loss on novel-view semantic rendering samples, and $\lambda_{\text{sem}}$ controls the relative weight of the auxiliary rendering objective. We set $\lambda_{\text{sem}} = 1$ in all experiments, since the per-sample values of $\mathcal{L}_{\text{sem}}$ and $\mathcal{L}_{\text{QA}}$ are comparable in scale.

\paragraph{Model training.} 
We adopt InternVL3.5~\citep{wang2025internvl3} and Qwen3-VL~\citep{Qwen3-VL} as base models. The visual encoder and geometry encoder remain frozen. We train the geometry adapters, camera adapter, shared projector, and the semantic prediction head, and follow prior work~\citep{zhu2025teachlmm} to update only the MLP gating and up-projection layers in the LLM decoder. This design enables efficient training while preserving the model's pretrained capabilities.
\section{Experiments}
% We compare our method with existing approaches in Sec.~\ref{sec:comparison} with 8B models, and then analyze how geometry features and the auxiliary novel-view semantic rendering task contribute to spatial reasoning in Sec.~\ref{sec:ablation} with 4B models. 

Our main experimental analysis centers on controlled experiments across two 4B base models, isolating the contributions of geometry features and auxiliary semantic supervision, as dense ablations at larger scale are computationally prohibitive. We additionally evaluate the complete training recipe with 8B backbones for system-level comparisons with existing methods, which differ in backbone, training data, and schedule.

\begin{figure}[t]
\centering
\includegraphics[width=1.0\textwidth]{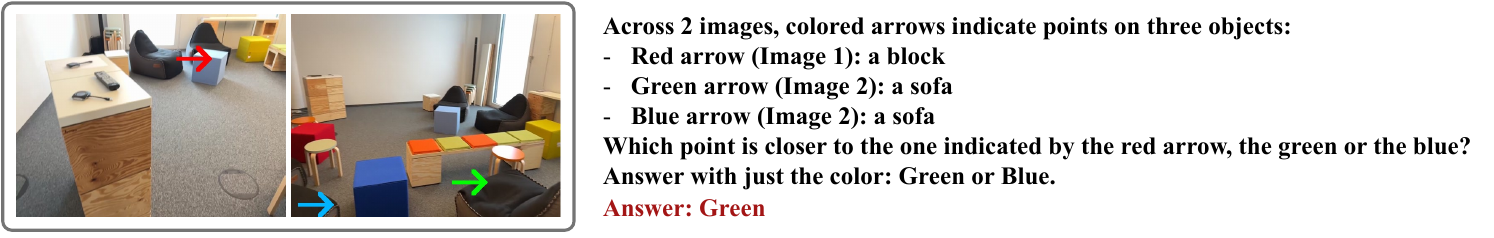}
\vspace{-1.7em}
\caption{
    % \textbf{Ambiguous Case.} When the second can is moved elsewhere and the third can is moved into the second can’s original location, our method only predicts that the third can has moved and cannot correctly infer the swap.
    \textbf{Visualization of P2P relative distance in 3D-Point-QA.} We render arrows as prompts to make recognition easier for VLMs~\citep{Xu2025VPBenchAC}. See more visualizations in Appendix \ref{app:3dpqa}.
}
\label{fig:low_level_vis}
\vspace{-1.4em}
\end{figure}

\subsection{Setup}
\label{sec:evalution_setup}
\paragraph{Benchmarks.}
We evaluate on VSI-Bench~\citep{Yang2024ThinkingIS}, ReVSI~\citep{zhang2026revsi}, and a set of low-level spatial QAs that we introduce. Existing spatial reasoning benchmarks (Sec.~\ref{sec:realted_3dvlm_dataset}) focus on high-level, multi-hop tasks where objects are referred to only by name. This makes it difficult to disentangle whether errors stem from geometric reasoning or from object grounding given text alone. 
SPAR~\citep{Zhang2025FromFT} extracts low-level spatial QAs but still relies on object centroids for answering, coupling geometric reasoning with object grounding.
We instead construct 3D-Point-QA, a dataset with low-level reasoning tasks (with visual and textual point prompts) that should be straightforward to answer given camera poses and point clouds without object-level reasoning. The dataset includes point-to-camera distance, point-to-point distance, relative distance comparison, point matching, and 3D coordinate mapping (Fig.~\ref{fig:low_level_vis} and Fig.~\ref{fig:low_level_qa_supp}). We extract 100K training and 5K validation QAs from ground-truth meshes in ScanNet++~\citep{Yeshwanth2023ScanNetAH} following the official splits.

\paragraph{Training data composition.}
Our training data consists of spatial QA pairs and semantic rendering samples constructed from indoor 3D datasets, including ScanNet, ScanNet++, and ARKitScenes. We use 494K total samples: 290K QA pairs from VSI-590K~\citep{Yang2025CambrianSTS}, 4K route-planning QA pairs from VLM-3R~\citep{Fan2025VLM3RVM}, 100K 3D-Point-QA pairs, and 100K semantic rendering pairs.

\paragraph{Metrics.}
For VSI-Bench and ReVSI, we follow prior work~\citep{Yang2024ThinkingIS} and report accuracy for multiple-choice questions and mean relative accuracy for numerical questions. For 3D-Point-QA, we use $\delta_{1.25}$~\citep{Eigen2014DepthMP} for point-to-camera and point-to-point distance estimation, accuracy for relative distance inference, PCK@0.05~\citep{Yang2013ArticulatedHD} for point matching, and the percentage of predicted points within 0.5m of the ground truth for 3D mapping.

\subsection{Comparisons with Existing Methods}
\label{sec:comparison}
We compare with existing methods on ReVSI~\citep{zhang2026revsi} and VSI-Bench~\citep{Yang2024ThinkingIS}.
Our method outperforms prior open-source methods while using less annotated training data.

\begin{table}[t]
    \centering
    \vspace{-0.3em}
    \caption{
    \textbf{VSI-Bench sub-task breakdown.}
    Best results within each model group are \textbf{bolded}. \textit{\#QA} denotes the number of spatial training QAs. 9B parameters include the frozen geometry encoder.
    }
    % \vspace{-0.6em}
    \resizebox{\textwidth}{!}{
            \begin{tabular}{l|lcr|c|cccccccc} & & & & & \rotatebox{75}{Obj. Count} & \rotatebox{75}{Abs. Dist.} & \rotatebox{75}{Obj. Size} & \rotatebox{75}{Room Size} & \rotatebox{75}{Rel. Dist.} & \rotatebox{75}{Rel. Dir.} & \rotatebox{75}{Route Plan} & \rotatebox{75}{Appr. Order} \\
            Methods & \multicolumn{1}{c}{Backbone} & \multicolumn{1}{c}{\#Params} & \multicolumn{1}{c|}{\#QA} & Avg. & \multicolumn{4}{c}{\cellcolor{orange!10}Numerical Answer} & \multicolumn{4}{c}{\cellcolor{yellow!10}Multiple-Choice Answer} \\
            \midrule
            \rowcolor{navyblue!5} \multicolumn{13}{l}{\textit{Baseline}} \\
            % Chance (Random) & -- & -- & -- & -- & -- & -- & -- & -- & 25.0 & 36.1 & 28.3 & 25.0 \\
            Chance (Frequency) & -- & -- & -- & 34.0 & 62.1 & 32.0 & 29.9 & 33.1 & 25.1 & 47.9 & 28.4 & 25.2 \\
            % Human & -- & -- & -- & 79.2 & 94.3 & 47.0 & 60.4 & 45.9 & 94.7 & 95.8 & 95.8 & 100.0 \\
            \hline
            \rowcolor{navyblue!5} \multicolumn{13}{l}{\textit{Proprietary Models (API)}} \\
            GPT-4o & -- & -- & -- & 34.0 & 46.2 & 5.3 & 43.8 & 38.2 & 37.0 & 41.3 & 31.5 & 28.5 \\
            Gemini-2.5 Pro & -- & -- & -- & 51.5 & 43.8 & 34.9 & 64.3 & 42.8 & 61.1 & 47.8 & 45.9 & 71.3 \\
            \hline \rowcolor{navyblue!5} \multicolumn{13}{l}{\textit{Open-source Fine-tuned Models}} \\
            VLM-3R & LLaVA-NeXT & 7B & 208K & 60.9 & 70.2 & 49.4 & 69.2 & 67.1 & 65.4 & 80.5 & 45.4 & 40.1 \\
            VG-LLM & Qwen2.5-VL & 9B & 442K & 62.2 & 71.4 & 56.8 & 69.0 & 69.1 & 67.9 & 83.2 & 47.4 & 32.5 \\
            % SenseNova-SI & InternVL3 & 8B & 800K & 60.9 & -- & -- & -- & -- & -- & -- & -- & -- \\
            SenseNova-SI & InternVL3 & 8B & 8M & 68.8 & 72.0 & 53.5 & 76.8 & 72.8 & 69.6 & 80.8 & 48.5 & 76.4 \\
            GeoThinker & Qwen3-VL & 9B & 1.1M & 72.6 & -- & -- & -- & -- & -- & -- & -- & -- \\
            Cambrian-S & Qwen2.5 & 7B & 590K & 67.5 & 73.2 & 50.5 & 74.9 & 72.2 & 71.1 & 76.2 & 41.8 & 80.1 \\
            Cambrian-P & Cambrian-S & 7B & 590K & 73.7 & \textbf{74.9} & 60.1 & 76.0 & \textbf{76.9} & 74.8 & 89.5 & \textbf{52.6} & 85.0 \\
            \rowcolor{oursrow}
            \textbf{Render2Reason} & InternVL3.5 & 9B & 394K & 67.3 & 72.0 & 51.9 & 71.8 & 68.2 & 68.2 & 83.5 & 45.9 & 76.4 \\ 
            \rowcolor{oursrow}\textbf{Render2Reason} & Qwen3-VL & 9B & 394K & \textbf{75.3} & 73.4 & \textbf{65.8} & \textbf{79.3} & 76.5 & \textbf{79.0} & \textbf{94.0} & 47.4 & \textbf{87.1} \\ 
            \bottomrule
        \end{tabular}
        }
    \label{tab:vsi_bench_main}
    \vspace{-1.3em}
\end{table}
\paragraph{Results on VSI-Bench.} 
We evaluate our method with 128 input views. In Tab.~\ref{tab:vsi_bench_main}, R2R with the Qwen3-VL backbone achieves 75.3 average accuracy, the best among all compared methods, surpassing Cambrian-P-7B (73.7) and GeoThinker-9B (72.6) with less training data. 
With the InternVL3.5 backbone, R2R reaches 67.3, surpassing VLM-3R-7B (60.9) and VG-LLM-9B (62.2) and on par with Cambrian-S-7B (67.5).
Across categories, our Qwen3-VL variant achieves the best results on five of eight question types. The largest margins over Cambrian-P are on absolute distance (+5.7), relative direction (+4.5), and relative distance (+4.2), and the absolute-distance and relative-direction gains are consistent with our results on ReVSI.
Overall, these results show that our auxiliary novel-view semantic rendering objective is an effective and data-efficient approach to improving spatial reasoning in VLMs.

\paragraph{Results on ReVSI.} 
ReVSI evaluates spatial reasoning across seven question types with manually refined annotations conditioned on the input frame count. We evaluate our method with 64 input frames and compare all methods under the same frame setting in Tab.~\ref{tab:revsi_main}.
R2R with the Qwen3-VL backbone reaches 54.5 average accuracy, the best among open-source models. It exceeds Cambrian-P by 2.5 points, Cambrian-S by 5.4 points, and GPT-5.2 by 3.6 points, while the much larger Gemini 3 Pro remains ahead at 60.9. Our InternVL3.5 variant reaches 51.4, exceeding Cambrian-S by 2.3 points and GPT-5.2 by 0.5 points. 
\begin{table}[t]
    \centering
    \vspace{-0.2em}
    \caption{
    \textbf{ReVSI sub-task breakdown.}
    Best results within each section are \textbf{bolded}. \textit{\#QA} denotes the number of geometry QAs generated with ground truth. 9B parameters include the geometry encoder. 
    % $^\ast$ denotes results from us using provided checkpoints.
    }
    \vspace{-.5em}
    \resizebox{\textwidth}{!}{
            \begin{tabular}{l|lcr|c|ccccccc} & & & & & \rotatebox{75}{Obj. Count} & \rotatebox{75}{Abs. Dist.} & \rotatebox{75}{Obj. Size} & \rotatebox{75}{Room Size} & \rotatebox{75}{Rel. Dist.} & \rotatebox{75}{Rel. Dir.} & \rotatebox{75}{Route Plan} \\
            Methods & \multicolumn{1}{c}{Backbone} & \multicolumn{1}{c}{\#Params} & \multicolumn{1}{c|}{\#QA} & Avg. & \multicolumn{4}{c}{\cellcolor{orange!10}Numerical Answer} & \multicolumn{3}{c}{\cellcolor{yellow!10}Multiple-Choice Answer} \\
            \midrule
            \rowcolor{navyblue!5} \multicolumn{12}{l}{\textit{Baseline}} \\
            % Chance (Random) & -- & -- & -- & -- & -- & -- & -- & -- & 23.7 & 26.8 & 26.0 \\
            Chance (Frequency) & -- & -- & -- & 31.4 & 52.2 & 40.1 & 17.4 & 20.9 & 25.8 & 31.9 & 30.2 \\
            \hline \rowcolor{navyblue!5} \multicolumn{12}{l}{\textit{Proprietary Models (64+ Frames)}} \\
            GPT-5.2 & -- & -- & -- & 50.9 & 56.2 & 41.5 & 73.9 & 63.0 & 48.4 & 34.9 & 38.2 \\
            Gemini 3 Pro & -- & -- & -- & 60.9 & 60.1 & 54.7 & 79.3 & 51.9 & 68.1 & 56.0 & 56.4 \\
            \hline \rowcolor{navyblue!5} \multicolumn{12}{l}{\textit{Open-source Fine-tuned Models (64+ Frames)}} \\
            VST & Qwen2.5-VL & 7B & 4.2M & 46.4 & 35.4 & 52.6 & \textbf{67.9} & 47.2 & \textbf{49.2} & 36.9 & 35.4 \\
            VLM-3R & LLaVA-NeXT & 7B & 208K & 50.2 & 42.0 & 61.4 & 64.6 & 51.1 & 46.2 & 49.2 & 36.9 \\
            Cambrian-S & Qwen2.5 & 7B & 590K & 49.1 & \textbf{48.4} & 60.5 & 65.5 & 46.7 & 37.1 & 48.5 & 37.0 \\
            Cambrian-P & Cambrian-S & 7B & 590K & 52.0 & 41.4 & 68.3 & 66.7 & 45.9 & 40.2 & 48.7 & 53.1 \\
            \rowcolor{oursrow}\textbf{Render2Reason} & InternVL-3.5 & 9B & 394K & 51.4 & 38.4 & 60.0 & 60.5 & 52.3 & 42.9 & 50.3 & \textbf{55.2} \\
            \rowcolor{oursrow}\textbf{Render2Reason} & Qwen3-VL & 9B & 394K & \textbf{54.5} & 42.2 & \textbf{74.0} & 67.3 & \textbf{58.5} & 41.3 & \textbf{51.6} & 46.6 \\
            \bottomrule
        \end{tabular}
        }
    \label{tab:revsi_main}
    \vspace{-1.2em}
\end{table}
Across categories, R2R shows the largest gains on absolute distance (+5.7 over Cambrian-P) and room size (+7.4 over the strongest baseline, VLM-3R). However, it performs worse than Cambrian-S on object counting and worse than VST and VLM-3R on relative distance. This pattern suggests that our auxiliary supervision benefits scene-level geometry and layout reasoning more than object-centric estimation such as counting.

\subsection{Ablations and Auxiliary Task Analysis}
\label{sec:ablation}
% We provide ablation studies in Tab.~\ref{tab:ablation}, isolating the contributions of geometry features and the auxiliary novel-view semantic rendering task, comparing semantic against RGB supervision as the rendering target, different semantic grid resolutions, and evaluating the rendering quality of the auxiliary head. We also train and evaluate GeoSR~\citep{zhang2026make} in our setup for a fair comparison. 

We provide ablation studies in Tab.~\ref{tab:ablation} to isolate the contributions of geometry features and the auxiliary novel-view semantic rendering task, compare semantic and RGB supervision as the rendering target, vary the semantic grid resolution, and compare with GeoSR~\citep{zhang2026make}, which we retrain in our setup for a fair comparison. Finally, we evaluate the rendering quality of the auxiliary head.

% \input{table/ablation_study_mixed}
% \noindent \textbf{Effect of Geometry Features.}
% In InternVL variant (Tab.~\ref{tab:abl_internvl}), adding VGGT features yields a clear improvement on 3D-Point-QA (+1.7) and a moderate gain on ReVSI (+0.9), but only a marginal change on VSI-Bench (+0.1). This pattern is consistent with the design of 3D-Point-QA. Geometry features contribute most to the benchmark that isolates geometric reasoning, and contribute less on high-level multi-hop benchmarks where visual features and language priors already provide useful signal. The pattern also holds in Qwen-variant (Tab.~\ref{tab:abl_qwen}) experiments.

\noindent \textbf{Effect of Geometry Features.}
On InternVL (Tab.~\ref{tab:abl_internvl}), adding VGGT features improves 3D-Point-QA by +1.7 and ReVSI by +0.9, with only a marginal change on VSI-Bench (+0.1). This pattern is consistent with the design of 3D-Point-QA. Geometry features contribute most to the benchmark that isolates geometric reasoning, and less to high-level multi-hop benchmarks where visual features and language priors already provide useful signal. On Qwen (Tab.~\ref{tab:abl_qwen}), the gain is again largest on 3D-Point-QA (+4.9), with smaller gains on VSI-Bench (+1.0) and ReVSI (+0.8).

\begin{table}[b]
    \centering
    \vspace{-0.5em}
    \caption{
    \textbf{Ablation studies.}
    We report results on VSI-Bench and ReVSI (16 frames) and on 3D-Point-QA. \textit{Sem.} refers to semantic rendering. Best results are \textbf{bolded}.  In (e), all models include VGGT features; we disable the geometry pathway at inference. See Appendix~\ref{app:ablation} for full results.
    }
    \vspace{-0.5em}
    \footnotesize
    \setlength{\tabcolsep}{3pt}%
    % Sets \tabrowht to the height of the tallest of \tabboxA/B/C,
    % so shorter panels in a row are vertically centered within it.
    \def\settabrowht{%
        \setlength{\tabrowht}{\dimexpr\ht\tabboxA+\dp\tabboxA\relax}%
        \ifdim\dimexpr\ht\tabboxB+\dp\tabboxB\relax>\tabrowht
            \setlength{\tabrowht}{\dimexpr\ht\tabboxB+\dp\tabboxB\relax}\fi
        \ifdim\dimexpr\ht\tabboxC+\dp\tabboxC\relax>\tabrowht
            \setlength{\tabrowht}{\dimexpr\ht\tabboxC+\dp\tabboxC\relax}\fi}%
    % ---------------- Row 1: (a) (b) (c) (no blank lines inside!) ----------------
    \sbox{\tabboxA}{%
        \begin{tabular}{l|ccc}
            \toprule
            Methods & VSI & ReVSI & 3DPQA \\
            \midrule
            Finetuned & 62.9 & 44.9 & 55.3 \\
            \quad + VGGT & 63.0 & 45.8 & 57.0 \\
            \rowcolor{oursrow} \quad + VGGT + Sem. & \textbf{64.6} & \textbf{48.0} & \textbf{59.9} \\
            \bottomrule
        \end{tabular}}%
    \sbox{\tabboxB}{%
        \begin{tabular}{l|ccc}
            \toprule
            Methods & VSI & ReVSI & 3DPQA \\
            \midrule
            Finetuned & 66.0 & 50.7 & 72.2 \\
            \quad + VGGT & 67.0 & 51.5 & 77.1 \\
            \rowcolor{oursrow} \quad + VGGT + Sem. & \textbf{68.2} & \textbf{52.5} & \textbf{80.0} \\
            \bottomrule
        \end{tabular}}%
    \sbox{\tabboxC}{%
        \begin{tabular}{l|ccc}
            \toprule
            Methods & VSI & ReVSI & 3DPQA \\
            \midrule
            Baseline & 67.0 & 51.5 & 77.1 \\
            GeoSR & 67.2 & 50.7 & \textbf{84.8} \\
            \rowcolor{oursrow} Ours & \textbf{68.2} & \textbf{52.5} & 80.0 \\
            \bottomrule
        \end{tabular}}%
    \settabrowht
    \begin{adjustbox}{max width=\textwidth}%
    \begin{subtable}[t]{\wd\tabboxA}
        \centering
        \caption{Components (InternVL).}
        \parbox[b][\tabrowht][c]{\linewidth}{\centering\usebox{\tabboxA}}
        \label{tab:abl_internvl}
    \end{subtable}%
    \hspace{1.5em}%
    \begin{subtable}[t]{\wd\tabboxB}
        \centering
        \caption{Components (Qwen).}
        \parbox[b][\tabrowht][c]{\linewidth}{\centering\usebox{\tabboxB}}
        \label{tab:abl_qwen}
    \end{subtable}%
    \hspace{1.5em}%
    \begin{subtable}[t]{\wd\tabboxC}
        \centering
        \caption{GeoSR comparison (Qwen).}
        \parbox[b][\tabrowht][c]{\linewidth}{\centering\usebox{\tabboxC}}
        \label{tab:abl_geosr}
    \end{subtable}%
    \end{adjustbox}

    \vspace{0.8em}

    % ---------------- Row 2: (d) (e) (f) (no blank lines inside!) ----------------
    \sbox{\tabboxA}{%
        \begin{tabular}{lc|ccc}
            \toprule
            Target & VGGT & VSI & ReVSI & 3DPQA \\
            \midrule
            Sem.       & \xmark & 61.6 & 45.9 & 60.6 \\
            RGB        & \cmark & 62.4 & 45.2 & 60.2 \\
            RGB + Sem. & \cmark & 62.6 & 46.6 & \textbf{61.1} \\
            \rowcolor{oursrow} Sem. & \cmark & \textbf{64.6} & \textbf{48.0} & 59.9 \\
            \bottomrule
        \end{tabular}}%
    \sbox{\tabboxB}{%
        \begin{tabular}{l|ccc}
            \toprule
            Grid & VSI & ReVSI & 3DPQA \\
            \midrule
            $8 \times 8$ & \textbf{64.9} & 46.2 & 53.0 \\
            \rowcolor{oursrow} $16 \times 16$ & 64.6 & 48.0 & \textbf{59.9} \\
            $32 \times 32$ & 64.3 & \textbf{48.4} & 56.6 \\
            \bottomrule
        \end{tabular}}%
    \sbox{\tabboxC}{%
        \begin{tabular}{l|ccc}
            \toprule
            Methods & VSI & ReVSI & 3DPQA \\
            \midrule
            InternVL & $-$6.3 & $-$5.5 & $-$6.4 \\
            \rowcolor{oursrow} \quad + Sem. & $-$6.7 & $-$7.3 & $-$7.0 \\
            Qwen & $-$2.3 & $-$5.2 & $-$5.2 \\
            \rowcolor{oursrow} \quad + Sem. & $-$4.2 & $-$8.9 & $-$8.7 \\
            \bottomrule
        \end{tabular}}%
    \settabrowht
    \begin{adjustbox}{max width=\textwidth}%
    \begin{subtable}[t]{\wd\tabboxA}
        \centering
        \caption{Rendering target (InternVL).}
        \parbox[b][\tabrowht][c]{\linewidth}{\centering\usebox{\tabboxA}}
        \label{tab:abl_target}
    \end{subtable}%
    \hspace{1.5em}%
    \begin{subtable}[t]{\wd\tabboxC}
        \centering
        \caption{Geometry removal in inference.}
        \parbox[b][\tabrowht][c]{\linewidth}{\centering\usebox{\tabboxC}}
        \label{tab:abl_alpha}
    \end{subtable}%
    \hspace{1.5em}%
    \begin{subtable}[t]{\wd\tabboxB}
        \centering
        \caption{Grid resolution (InternVL).}
        \parbox[b][\tabrowht][c]{\linewidth}{\centering\usebox{\tabboxB}}
        \label{tab:abl_grid}
    \end{subtable}%
    \end{adjustbox}
    \vspace{-1em}
    \label{tab:ablation}
\end{table}

\noindent \textbf{Effect of Auxiliary Task.}
Adding novel-view semantic rendering on top of VGGT features improves all three benchmarks on both backbones (+1.6, +2.2, and +2.9 on VSI-Bench, ReVSI, and 3D-Point-QA for InternVL; +1.2, +1.0, and +2.9 for Qwen). As shown in the per-category results (Appendix~\ref{app:ablation}), both backbones gain on point-to-camera distance, point-to-point distance, and 3D mapping in 3D-Point-QA, and on relative distance in both high-level benchmarks, while room size in VSI-Bench drops and other categories vary. Without VGGT (Tab.~\ref{tab:abl_target}), the auxiliary task alone improves 3D-Point-QA and ReVSI but hurts VSI-Bench ($-$1.3), while VGGT alone yields only marginal gains, indicating the two are complementary.

\noindent \textbf{Semantic vs. RGB Supervision.}
We compare semantic-class prediction against patch-level RGB prediction as the rendering target with the same architecture and VGGT inputs. RGB supervision improves low-level metrics, particularly point matching, where patch-level reconstruction directly aids dense correspondence. 
However, on high-level reasoning (VSI-Bench, ReVSI), RGB supervision underperforms semantic supervision, likely because appearance-level prediction does not tie geometry to object-level semantics, which is important for multi-hop spatial questions. 
Combining both supervisions yields the best 3D-Point-QA score but degrades multi-hop performance relative to semantic supervision alone. 
We therefore use semantic rendering as our default supervision target.

% \noindent \textbf{Perofrmance without geometry during infernece.}
% To better understand whether models use more geometry feature, we disable the geometry pathway at inference by setting the fusion gate $\alpha{=}0$ and measure the change in performance (Tab.~\ref{tab:abl_alpha}). Since both models are trained with geometry features, removing them is off-distribution for both, and both fall below the finetuned baseline without VGGT. We therefore compare the size of the drop between the two models rather than its absolute value. On both backbones and all three benchmarks, removing geometry hurts the auxiliary-task model more than the geometry-augmented baseline. On Qwen, the auxiliary-task model falls slightly below the baseline once geometry is removed ($-$0.8, $-$2.7, and $-$0.6 on VSI-Bench, ReVSI, and 3D-Point-QA), indicating that its gains depend entirely on the geometry pathway. On InternVL, most of the ReVSI gain disappears (from +2.2 to +0.4), while most of the VSI-Bench (from +1.6 to +1.2) and 3D-Point-QA (from +2.9 to +2.3) gains persist, suggesting that on this backbone the auxiliary task also improves performance through routes other than geometry integration.

\noindent \textbf{Performance without Geometry at Inference.}
To test whether the auxiliary task increases reliance on geometry, we disable the geometry pathway at inference ($\alpha{=}0$) and compare the performance drops (Tab.~\ref{tab:abl_alpha}). This is off-distribution for both models (both fall below the finetuned baseline), so only the difference between their drops is meaningful. On both backbones and all three benchmarks, removing geometry hurts the auxiliary-task models more, especially on Qwen, suggesting that our auxiliary task encourages the model to rely more on geometry features.

% \noindent \textbf{Semantic Grid Resolution.}
% As shown in Tab.~\ref{tab:abl_grid}, the $16 \times 16$ grid gives the best overall balance. Both coarser ($8 \times 8$) and finer ($32 \times 32$) grids perform comparably on the high-level benchmarks but drop on 3D-Point-QA ($-$6.9 and $-$3.3).

\noindent \textbf{Comparison with GeoSR.}
Under the same setup (Tab.~\ref{tab:abl_geosr}), GeoSR achieves a higher 3D-Point-QA score (84.8 vs.\ 80.0), as its appearance masking directly strengthens the geometry readout that 3D-Point-QA measures. However, our method outperforms GeoSR on VSI-Bench (+1.0) and ReVSI (+1.8), since multi-hop spatial reasoning requires both geometric and object-level semantic understanding, which our semantic rendering task provides.

\noindent \textbf{Semantic Grid Resolution.}
As shown in Tab.~\ref{tab:abl_grid}, the $16 \times 16$ grid gives the best overall balance. Both coarser and finer grids perform comparably on the high-level benchmarks but drop on 3D-Point-QA.

\begin{figure}[t]
\centering
\includegraphics[width=1.0\textwidth]{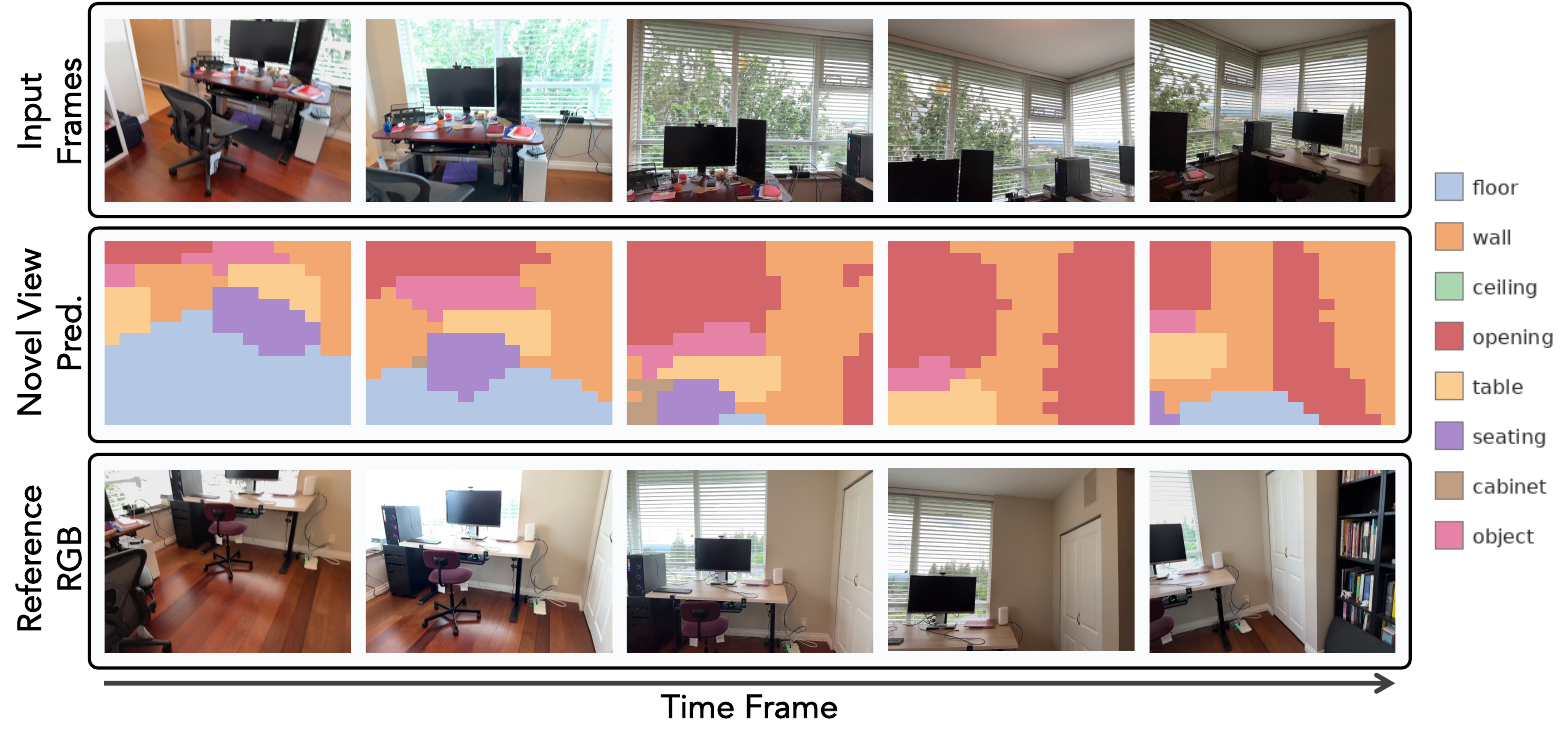}
\vspace{-2.em}
\caption{
    % \textbf{Ambiguous Case.} When the second can is moved elsewhere and the third can is moved into the second can’s original location, our method only predicts that the third can has moved and cannot correctly infer the swap.
    % \textbf{Semantic Rendering Visualization.} We visualize sequential novel-view semantic renderings in a forward-moving sequence. The first row shows the 5 input frames. Each subsequent prediction uses the previous 5 color images as input: the first prediction in row 2 uses all 5 frames from row 1; the second prediction uses frames 2-5 of row 1 plus the first image of row 2; and so on.
    \textbf{Semantic Rendering Visualization.} We visualize novel-view semantic renderings along a forward-moving camera trajectory. The first row shows the five initial RGB input frames. Each prediction uses the five RGB frames right before its target frame: the first prediction uses all five frames in row 1; the second uses frames 2–5 from row 1 and the first RGB frame in row 3; and so on.
}
\label{fig:semantic_vis}
\vspace{-1.1em}
\end{figure}

\noindent \textbf{Novel View Rendering Results.}
% We evaluate auxiliary semantic rendering on 5K held-out samples where the target view occurs 1 second after the last input frame. Our model achieves 44.7\% mIoU and 67.0\% patch accuracy, while a closest-input-frame copying baseline has 28.7\%mIoU and 50.1\%accuracy. These improvements indicate that the model’s predictions go beyond simply reusing a nearby observed semantic layout.
% Fig.~\ref{fig:semantic_vis} shows qualitative examples. The predicted semantic layouts are coherent with the target view, recovering room structure and object placement. Predicted videos are provided in the supplement.
We evaluate auxiliary semantic rendering on 5K held-out samples where the target view occurs 1 second after the last input frame. Our model (InternVL3.5-4B) achieves 44.7\% mIoU and 67.0\% patch accuracy. An oracle baseline that copies the best-matching input frame's semantic map for each target frame achieves 35.6\% mIoU and 57.6\% patch accuracy, indicating that the model accounts for viewpoint changes rather than copying an observed layout. Fig.~\ref{fig:semantic_vis} shows qualitative examples: the predicted layouts are largely consistent with the target view, capturing room structure and object placement. Predicted videos are provided in the supplement.

\vspace{-.7em}
\section{Conclusion}
\vspace{-.3em}
% We present R2R, a vision-language model for spatial reasoning. Standard spatial QA provides weak gradients for the geometric pathway in geometry-augmented VLMs, leaving it underutilized. To address this, we propose \textbf{novel-view semantic rendering} as an auxiliary training task that requires both visual and geometric pathways to predict the semantic layout of an unobserved viewpoint, and we show it increases the contribution of geometry features to spatial QA. We additionally introduce 3D-Point-QA, a low-level dataset that isolates geometric reasoning from object grounding. R2R achieves competitive performance on VSI-Bench, ReVSI, and 3D-Point-QA.
We present R2R, a vision-language model for spatial reasoning. To strengthen the supervision signal beyond standard spatial QA, we propose novel-view semantic rendering as an auxiliary training task that requires combining geometric and visual reasoning to predict the semantic layout of an unobserved viewpoint. We show that this auxiliary supervision consistently improves QA performance over a geometry-augmented baseline. We additionally introduce 3D-Point-QA, a low-level dataset that isolates geometric reasoning from object grounding. R2R achieves competitive performance on VSI-Bench, ReVSI, and 3D-Point-QA.

% \vspace{-0.5em}
\paragraph{Limitations.}
% Our auxiliary task can be supervised from any video data without annotation, but we validate it only at a limited scale (ScanNet~\citep{Dai2017ScanNetR3}, ScanNet++~\citep{Yeshwanth2023ScanNetAH}, ARKitScenes~\citep{Dehghan2021ARKitScenesAD}). Scaling to larger and more diverse scenes is a promising direction for future work.
Our auxiliary task can be supervised from any video data without human annotation, but we validate it only at a limited scale ~\citep{Dai2017ScanNetR3,Yeshwanth2023ScanNetAH,Dehghan2021ARKitScenesAD}. Scaling to larger and more diverse scenes is a promising direction for future work.

\paragraph{Acknowledgment.}
This work is supported in part by NSF IIS grant 2312102.
S.W. is supported by NSF 2331878 and 2340254, and research grants from Intel, Amazon, and IBM.
This research used both the DeltaAI advanced computing and data resource, which is supported by the National Science Foundation (award OAC 2320345) and the State of Illinois, and the Delta advanced computing and data resource which is supported by the National Science Foundation (award OAC 2005572) and the State of Illinois. Delta and DeltaAI are joint efforts of the University of Illinois Urbana-Champaign and its National Center for Supercomputing Applications.

% \input{tex/statements}

% \clearpage

{
    \small
    \bibliographystyle{iclr2027_conference}
    \bibliography{main}
}

%%%%%%%%%%%%%%%%%%%%%%%%%%%%%%%%%%%%%%%%%%%%%%%%%%%%%%%%%%%%
\clearpage
\appendix

\section{Appendix}

\subsection{Implementation Details}
\label{app:impl}

% \paragraph{Model.}
We adopt InternVL3.5~\citep{wang2025internvl3} and Qwen3-VL~\citep{Qwen3-VL} as base models. The visual encoder and geometry encoder remain frozen. We train the geometry adapters, camera adapter, shared projector, and semantic prediction head, and follow prior work~\citep{zhu2025teachlmm} to update only the MLP gating and up-projection layers in the LLM decoder. This enables efficient training while preserving the pretrained capabilities of the model. We use 16 input frames for the 4B variants and 32 for the 8B variants, but the 8B variants still perform well with more input frames at inference.

% \paragraph{Rendering heads and query tokens.}
For both semantic and RGB rendering, the prediction head is a two-layer MLP that outputs a $16\times16$ grid. Semantic rendering is supervised with cross-entropy over class indices, and RGB rendering with an MSE loss weighted by 20. We add new anchor tokens (e.g., \texttt{<SEM>}-style tokens) to the tokenizer vocabulary rather than repurposing existing tokens, and initialize their embeddings via per-dimension std sampling with row-norm renormalization. During development, we found that excluding these anchor tokens from the supervision signal yields better performance, so the auxiliary loss is computed only on the patch-level prediction tokens.

% \paragraph{Optimization.}
We train with AdamW~\citep{loshchilov2019adamw} using a learning rate of $1\times10^{-5}$, batch size 384, and zero weight decay, with a cosine schedule and 3\% linear warmup. We train for 1 epoch in all experiments, except Qwen3-VL-8B, which we train for 3 epochs as it converges more slowly. All ablations use identical training schedules.
Models are trained in bfloat16 with gradient checkpointing, using DeepSpeed ZeRO Stage 2~\citep{rajbhandari2020zero} on 8 NVIDIA GH200 GPUs. Training takes approximately 12 hours for the 4B variants, 30 hours for InternVL3.5-8B, and 96 hours for Qwen3-VL-8B.

% \paragraph{Compute.}

\subsection{Additional Rendering Results}
\begin{figure}[h]
\centering
\vspace{-.5em}
\includegraphics[width=0.91\textwidth]{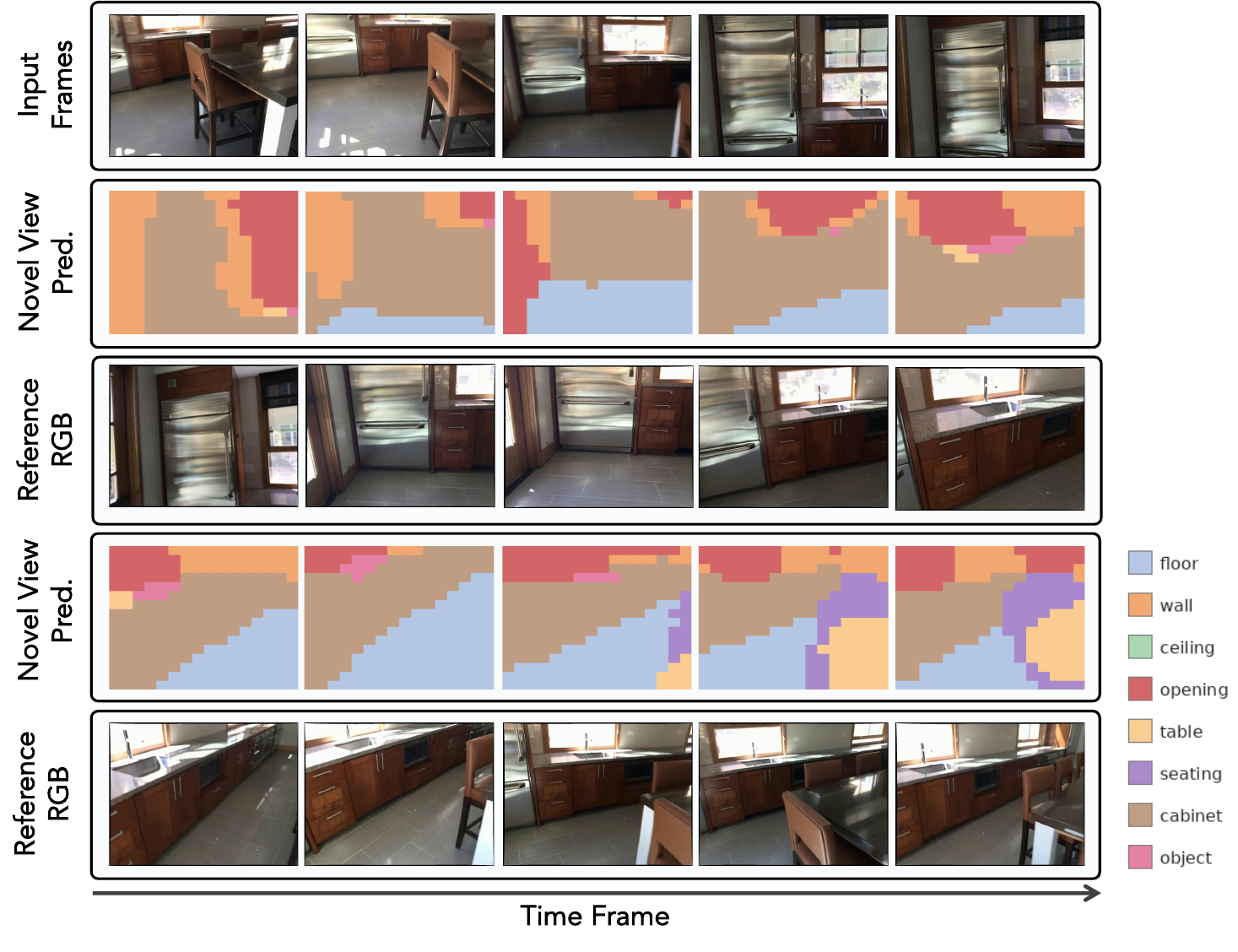}
\vspace{-1.5em}
\caption{
\textbf{Semantic Rendering Visualization.} We visualize sequential novel-view semantic renderings in a forward-moving sequence, with one-second intervals between adjacent views. The first row shows the 5 input frames. Each subsequent prediction uses the previous 5 frames as input (sliding window): the first prediction in row 2 uses all 5 frames from row 1; the second prediction uses frames 2-5 of row 1 plus the first RGB frame in row 3; and so on.}
\label{fig:semantic_vis_supp}
\vspace{-1.em}
\end{figure}

We visualize additional novel-view semantic rendering predictions in Fig.~\ref{fig:semantic_vis_supp}.

\subsection{Alternative Auxiliary Objectives}
\label{app:motivation}

As discussed in Sec.~\ref{sec:preliminaries}, standard QA training provides weak gradient signal through the geometry pathway. Before arriving at novel-view semantic rendering, we explored two alternatives to address this.

\paragraph{Geometry--visual alignment pretraining.}
We first tested whether the weak geometry signal stems from misalignment between geometry and visual features. We pretrained the geometry projector with geometry-only captioning and instruction following before joint spatial QA training. This improved geometry-only inference but not joint inference, indicating that better alignment alone does not lead the model to combine the two pathways.

\paragraph{Novel-view geometry feature prediction.}
We then trained the model to predict VGGT latent features at novel views, supervised through depth and camera poses decoded by the frozen VGGT heads. This reduced VSI-Bench accuracy by 1.5 points relative to the geometry-augmented baseline. We attribute this to the prediction target lying entirely in the geometry feature space, so it can be solved by the geometry pathway alone without engaging the visual pathway.

Both results suggest that an effective auxiliary objective should require both pathways. This motivates novel-view semantic rendering (Sec.~\ref{sec:semantic_rendering}), where localizing visible surfaces from the target pose requires geometry, and labeling them requires appearance.

% ===================================================================
% Appendix: full ablation results (per-subtask breakdown of Tab.~\ref{tab:ablation})
% Requires: booktabs, graphicx, colortbl (or xcolor with [table])
% "--" = placeholder to fill. Bold the best result per column after filling.
% ===================================================================

\subsection{Full Ablation Results}
\label{app:ablation}

We report the per-category results for the ablations in Tab.~\ref{tab:ablation}.
Tabs.~\ref{tab:app_abl_internvl_vsi} and~\ref{tab:app_abl_internvl_revsi} cover the component and rendering-target ablations on InternVL3.5-4B (Tabs.~\ref{tab:abl_internvl} and~\ref{tab:abl_target}). Tabs.~\ref{tab:app_abl_qwen_vsi} and~\ref{tab:app_abl_qwen_revsi} cover the component ablation and GeoSR comparison on Qwen3-VL-4B (Tabs.~\ref{tab:abl_qwen} and~\ref{tab:abl_geosr}). Tabs.~\ref{tab:app_abl_grid_vsi} and~\ref{tab:app_abl_grid_revsi} cover the grid resolution ablation (Tab.~\ref{tab:abl_grid}).

% ===================================================================
% 1. InternVL3.5-4B: components + rendering target  (panels a, c)
% ===================================================================
\begin{table}[h]
    \centering
    \caption{
    \textbf{Full results of the component and rendering-target ablations on VSI-Bench.}
    All models use InternVL3.5-4B with 16 input frames. Best results are \textbf{bolded}.
    }
    \resizebox{0.9\textwidth}{!}{
        \begin{tabular}{l|c|cccccccc}
        & & \rotatebox{75}{Obj. Count} & \rotatebox{75}{Abs. Dist.} & \rotatebox{75}{Obj. Size} & \rotatebox{75}{Room Size} & \rotatebox{75}{Rel. Dist.} & \rotatebox{75}{Rel. Dir.} & \rotatebox{75}{Route Plan} & \rotatebox{75}{Appr. Order} \\
        Methods & Avg. & \multicolumn{4}{c}{\cellcolor{orange!10}Numerical Answer} & \multicolumn{4}{c}{\cellcolor{yellow!10}Multiple-Choice Answer} \\
        \midrule
        Finetuned & 62.9 & \textbf{70.6} & 45.5 & \textbf{70.2} & 62.6 & 59.3 & 70.8 & 40.7 & \textbf{83.3} \\
        \quad + VGGT & 63.0 & 69.7 & \textbf{47.6} & 67.8 & \textbf{66.3} & 58.9 & 75.5 & 42.8 & 75.2 \\
        \quad + Sem. & 61.6 & 69.3 & 41.3 & 69.5 & 63.4 & 61.0 & 66.0 & 40.7 & 81.6 \\
        \quad + VGGT + RGB & 62.4 & 69.7 & 42.7 & 69.2 & 64.2 & 61.0 & 72.2 & 45.4 & 74.9 \\
        \quad + VGGT + RGB + Sem. & 62.6 & 69.0 & 42.8 & 68.1 & 65.5 & 60.1 & 74.0 & 44.3 & 77.0 \\
        \rowcolor{oursrow}\textbf{\quad + VGGT + Sem.} & \textbf{64.6} & 70.4 & 44.7 & 69.3 & 64.6 & \textbf{61.8} & \textbf{78.3} & \textbf{47.9} & 79.6 \\
        \bottomrule
    \end{tabular}}
\label{tab:app_abl_internvl_vsi}
\end{table}

\begin{table}[h]
    \centering
    \caption{
    \textbf{Full results of the component and rendering-target ablations on ReVSI-16-Frame and 3D-Point-QA.}
    All models use InternVL3.5-4B. Best results are \textbf{bolded}.
    }
    \resizebox{\textwidth}{!}{
        \begin{tabular}{l|cccccc|cccccc}
        & \multicolumn{6}{c|}{ReVSI-16-Frame} & \multicolumn{6}{c}{3D-Point-QA} \\
        Methods & Avg. & \rotatebox{75}{Obj. Cnt.} & \rotatebox{75}{Abs. Dist.} & \rotatebox{75}{Obj. Size} & \rotatebox{75}{Rel. Dist.} & \rotatebox{75}{Rel. Dir.} & Avg. & \rotatebox{75}{P2C Dist.} & \rotatebox{75}{P2P Dist.} & \rotatebox{75}{P2P R-Dist.} & \rotatebox{75}{Point Match} & \rotatebox{75}{3D Mapping} \\
        \midrule
        Finetuned & 44.9 & 33.6 & 53.4 & 57.7 & 35.0 & 45.0 & 55.3 & 66.0 & 41.2 & 90.1 & 34.8 & 44.7 \\
        \quad + VGGT & 45.8 & 32.3 & 55.0 & 57.0 & 36.4 & 48.1 & 57.0 & 59.6 & 44.0 & 92.5 & 44.1 & 45.0 \\
        \quad + Sem. & 45.9 & 34.6 & 51.7 & 58.4 & 36.1 & 48.5 & 60.6 & 69.5 & \textbf{54.6} & 91.3 & 38.2 & 49.3 \\
        \quad + VGGT + RGB & 45.2 & 33.1 & 50.5 & 58.1 & 35.3 & 49.1 & 60.2 & 65.3 & 44.4 & 92.1 & \textbf{48.6} & \textbf{50.7} \\
        \quad + VGGT + RGB + Sem. & 46.6 & 35.4 & 55.0 & \textbf{58.7} & 36.1 & 47.9 & \textbf{61.1} & \textbf{70.3} & 52.9 & \textbf{93.3} & 38.5 & 50.6 \\
        \rowcolor{oursrow}\textbf{\quad + VGGT + Sem.} & \textbf{48.0} & \textbf{37.4} & \textbf{55.6} & 57.1 & \textbf{38.4} & \textbf{51.6} & 59.9 & 67.9 & 52.9 & 91.3 & 38.3 & 49.2 \\
        \bottomrule
    \end{tabular}}
\label{tab:app_abl_internvl_revsi}
\end{table}

% ===================================================================
% 2. Qwen3-VL-4B: components + GeoSR comparison  (panels b, e)
%    "+ VGGT" is the Baseline row of panel (e).
% ===================================================================
\begin{table}[h]
    \centering
    \caption{
    \textbf{Full results of the component ablation and GeoSR comparison on VSI-Bench.}
    All models use Qwen3-VL-4B with 16 input frames. Best results are \textbf{bolded}.
    }
    \resizebox{0.9\textwidth}{!}{
        \begin{tabular}{l|c|cccccccc}
        & & \rotatebox{75}{Obj. Count} & \rotatebox{75}{Abs. Dist.} & \rotatebox{75}{Obj. Size} & \rotatebox{75}{Room Size} & \rotatebox{75}{Rel. Dist.} & \rotatebox{75}{Rel. Dir.} & \rotatebox{75}{Route Plan} & \rotatebox{75}{Appr. Order} \\
        Methods & Avg. & \multicolumn{4}{c}{\cellcolor{orange!10}Numerical Answer} & \multicolumn{4}{c}{\cellcolor{yellow!10}Multiple-Choice Answer} \\
        \midrule
        Finetuned & 66.0 & 70.7 & 47.0 & \textbf{75.8} & \textbf{71.8} & 64.9 & 71.1 & 43.8 & 83.2 \\
        \quad + VGGT & 67.0 & 68.8 & 51.6 & 73.6 & 71.6 & 63.0 & 79.3 & 43.3 & 85.0 \\
        \quad + VGGT + GeoSR & 67.2 & \textbf{71.4} & 53.7 & 73.0 & 63.2 & 67.5 & \textbf{80.8} & \textbf{44.3} & 84.0 \\
        \rowcolor{oursrow}\textbf{\quad + VGGT + Sem.} & \textbf{68.2} & 70.6 & \textbf{56.0} & 75.7 & 65.8 & \textbf{72.8} & 77.5 & 41.2 & \textbf{85.6} \\
        \bottomrule
    \end{tabular}}
\label{tab:app_abl_qwen_vsi}
\end{table}

\begin{table}[h]
    \centering
    \caption{
    \textbf{Full results of the component ablation and GeoSR comparison on ReVSI-16-Frame and 3D-Point-QA.}
    All models use Qwen3-VL-4B. Best results are \textbf{bolded}.
    }
    \resizebox{\textwidth}{!}{
        \begin{tabular}{l|cccccc|cccccc}
        & \multicolumn{6}{c|}{ReVSI-16-Frame} & \multicolumn{6}{c}{3D-Point-QA} \\
        Methods & Avg. & \rotatebox{75}{Obj. Cnt.} & \rotatebox{75}{Abs. Dist.} & \rotatebox{75}{Obj. Size} & \rotatebox{75}{Rel. Dist.} & \rotatebox{75}{Rel. Dir.} & Avg. & \rotatebox{75}{P2C Dist.} & \rotatebox{75}{P2P Dist.} & \rotatebox{75}{P2P R-Dist.} & \rotatebox{75}{Point Match} & \rotatebox{75}{3D Mapping} \\
        \midrule
        Finetuned & 50.7 & 38.6 & 59.8 & \textbf{68.4} & 37.8 & 49.0 & 72.2 & 81.7 & 63.1 & 92.7 & 64.5 & 58.9 \\
        \quad + VGGT & 51.5 & \textbf{39.2} & 65.1 & 64.3 & 38.2 & \textbf{50.4} & 77.1 & 82.5 & 68.6 & 95.6 & 69.9 & 68.9 \\
        \quad + VGGT + GeoSR & 50.7 & 39.0 & 65.4 & 63.1 & \textbf{39.6} & 46.6 & \textbf{84.8} & \textbf{90.2} & \textbf{78.8} & \textbf{97.0} & \textbf{75.1} & \textbf{83.0} \\
        \rowcolor{oursrow}\textbf{\quad + VGGT + Sem.} & \textbf{52.5} & 39.0 & \textbf{72.6} & 62.8 & 39.1 & 48.8 & 80.0 & 85.3 & 69.9 & 96.6 & 72.6 & 75.7 \\
        \bottomrule
    \end{tabular}}
\label{tab:app_abl_qwen_revsi}
\end{table}

% ===================================================================
% 3. Grid resolution, InternVL3.5-4B  (panel d)
% ===================================================================
\begin{table}[h]
    \centering
    \caption{
    \textbf{Full results of the grid resolution ablation on VSI-Bench.}
    All models use InternVL3.5-4B with VGGT features, semantic rendering, and 16 input frames. Best results are \textbf{bolded}.
    }
    \resizebox{0.8\textwidth}{!}{
        \begin{tabular}{l|c|cccccccc}
        & & \rotatebox{75}{Obj. Count} & \rotatebox{75}{Abs. Dist.} & \rotatebox{75}{Obj. Size} & \rotatebox{75}{Room Size} & \rotatebox{75}{Rel. Dist.} & \rotatebox{75}{Rel. Dir.} & \rotatebox{75}{Route Plan} & \rotatebox{75}{Appr. Order} \\
        Grid & Avg. & \multicolumn{4}{c}{\cellcolor{orange!10}Numerical Answer} & \multicolumn{4}{c}{\cellcolor{yellow!10}Multiple-Choice Answer} \\
        \midrule
        $8\times8$ & \textbf{64.9} & 69.5 & \textbf{47.0} & 68.0 & \textbf{64.9} & \textbf{64.8} & 75.7 & \textbf{48.5} & \textbf{80.7} \\
        \rowcolor{oursrow}\textbf{$16\times16$} & 64.6 & 70.4 & 44.7 & \textbf{69.3} & 64.6 & 61.8 & \textbf{78.3} & 47.9 & 79.6 \\
        $32\times32$ & 64.3 & \textbf{70.6} & 46.3 & 69.2 & 61.7 & 63.4 & 77.0 & 45.9 & 80.6 \\
        \bottomrule
    \end{tabular}}
\label{tab:app_abl_grid_vsi}
\end{table}

\begin{table}[h]
    \centering
    \caption{
    \textbf{Full results of the grid resolution ablation on ReVSI-16-Frame and 3D-Point-QA.}
    All models use InternVL3.5-4B with VGGT features and semantic rendering. Best results are \textbf{bolded}.
    }
    \resizebox{\textwidth}{!}{
        \begin{tabular}{l|cccccc|cccccc}
        & \multicolumn{6}{c|}{ReVSI-16-Frame} & \multicolumn{6}{c}{3D-Point-QA} \\
        Grid & Avg. & \rotatebox{75}{Obj. Cnt.} & \rotatebox{75}{Abs. Dist.} & \rotatebox{75}{Obj. Size} & \rotatebox{75}{Rel. Dist.} & \rotatebox{75}{Rel. Dir.} & Avg. & \rotatebox{75}{P2C Dist.} & \rotatebox{75}{P2P Dist.} & \rotatebox{75}{P2P R-Dist.} & \rotatebox{75}{Point Match} & \rotatebox{75}{3D Mapping} \\
        \midrule
        $8\times8$ & 46.2 & 34.2 & 55.4 & 55.3 & \textbf{38.7} & 47.7 & 53.0 & 65.2 & 44.5 & 85.0 & 26.4 & 43.9 \\
        \rowcolor{oursrow}\textbf{$16\times16$} & 48.0 & \textbf{37.4} & 55.6 & 57.1 & 38.4 & \textbf{51.6} & \textbf{59.9} & 67.9 & \textbf{52.9} & \textbf{91.3} & \textbf{38.3} & \textbf{49.2} \\
        $32\times32$ & \textbf{48.4} & 36.4 & \textbf{60.5} & \textbf{59.4} & 38.4 & 47.4 & 56.6 & \textbf{71.5} & 51.2 & 86.8 & 29.1 & 44.1 \\
        \bottomrule
    \end{tabular}}
\label{tab:app_abl_grid_revsi}
\end{table}

\clearpage
\subsection{Target Frame Information Leakage Analysis}
\label{app:leakage}
For novel-view semantic rendering, we check whether target-frame information leaks into the context-view features through VGGT’s global attention, by recomputing the context-view features without the target frame at inference, while keeping the target camera token unchanged. On the 5K held-out rendering samples, only 2.3\% of predicted patches change, and performance is nearly identical (44.3\% vs.\ 44.7\% mIoU; 66.7\% vs.\ 67.0\% patch accuracy). This indicates that the model does not rely on target-frame information leaked through the context features.

\subsection{3D-Point-QA Visualization}
\label{app:3dpqa}
\begin{figure}[h]
\centering
\includegraphics[width=1.0\textwidth]{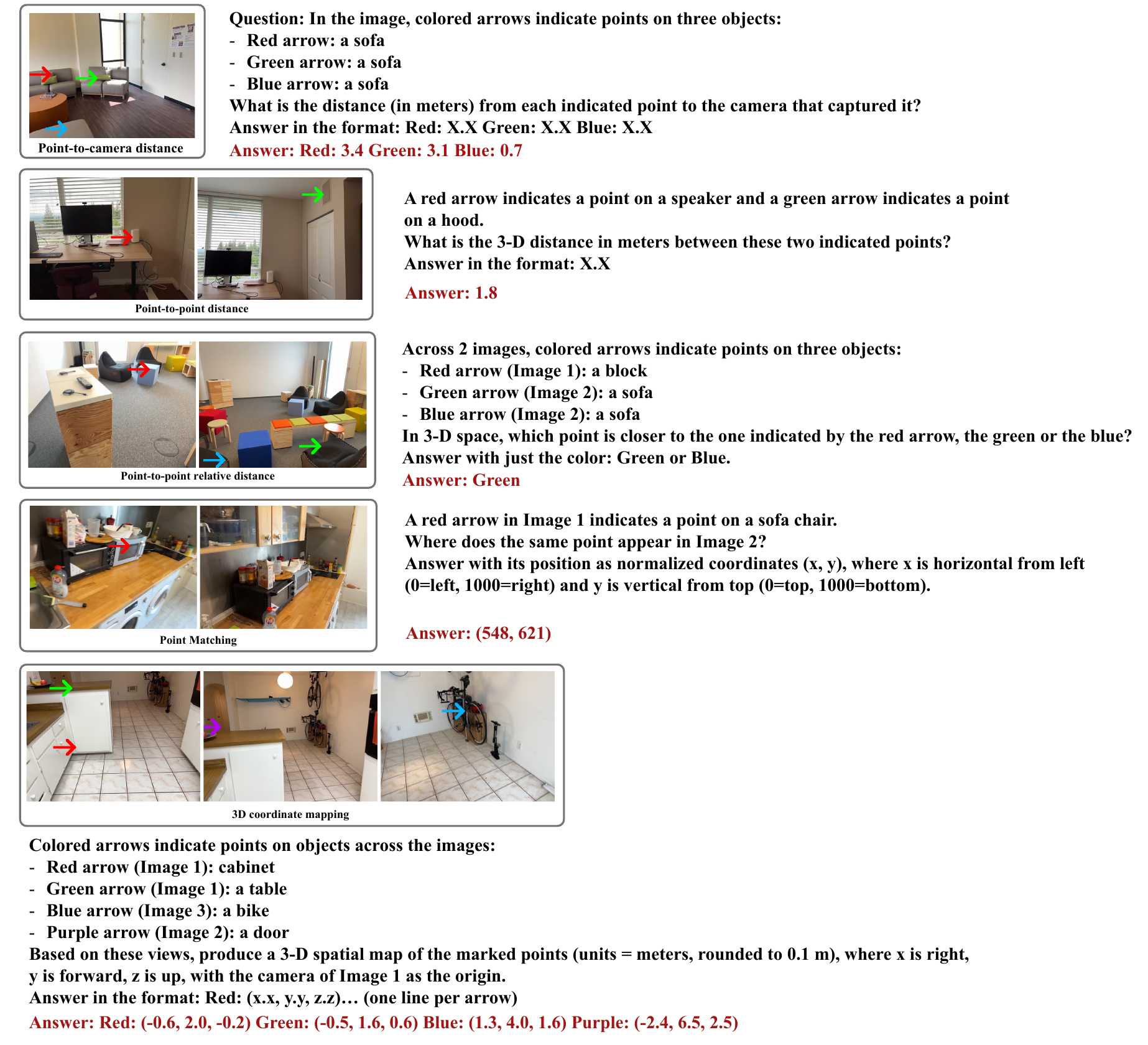}
\vspace{-1.7em}
\caption{
    % \textbf{Ambiguous Case.} When the second can is moved elsewhere and the third can is moved into the second can’s original location, our method only predicts that the third can has moved and cannot correctly infer the swap.
    \textbf{Visualization of 3D-Point-QA examples.} We visualize the question and answer for each sample.
}
\label{fig:low_level_qa_supp}
\vspace{-1em}
\end{figure}

We visualize the full question prompts alongside the corresponding inputs in Fig.~\ref{fig:low_level_qa_supp}. Note that the arrows are enlarged here for clarity. The actual inputs use smaller arrows, following the size convention from DepthLM~\citep{cai2025depthlm}.

\subsection{Zero-Shot Evaluation on MindCube}
\label{sec:mindcube}

To test whether the benefit of novel-view semantic rendering transfers beyond the benchmarks used during development, we evaluate zero-shot on MindCube-Tiny~\citep{wang2025spatial}, a benchmark for spatial mental modeling from limited views. We compare the original pretrained InternVL3.5-4B with the three variants from Tab.~\ref{tab:abl_internvl} (Finetuned, + VGGT, and + VGGT + Sem.), using the same checkpoints without any additional training. MindCube-Tiny groups questions into three settings by camera configuration: \emph{Rotation}, \emph{Among}, and \emph{Around}. Results are shown in Tab.~\ref{tab:mindcube}.

\begin{table}[h]
    \centering
    \caption{
    \textbf{Zero-shot results on MindCube-Tiny.}
    All models are based on InternVL3.5-4B and use the checkpoints from Tab.~\ref{tab:abl_internvl} without additional training.  Best results are \textbf{bolded}.
    }
    \begin{tabular}{l|ccc|c}
        \toprule
        Methods & Rotation & Among & Around & Overall \\
        \midrule
        Pretrained & \textbf{37.5} & 35.8 & 42.0 & 37.6 \\
        Finetuned & 30.5 & 39.3 & 48.4 & 39.8 \\
        \quad + VGGT & 37.0 & 38.2 & 34.4 & 37.1 \\
        \rowcolor{oursrow}\textbf{\quad + VGGT + Sem.} & 32.0 & \textbf{41.5} & \textbf{50.8} & \textbf{41.9} \\
        \bottomrule
    \end{tabular}
\label{tab:mindcube}
\end{table}

Adding geometry features alone hurts overall performance (39.8 $\rightarrow$ 37.1), most obviously on \emph{Around} (48.4 $\rightarrow$ 34.4). Adding novel-view semantic rendering makes the geometry features beneficial, achieving the best overall accuracy (41.9), a gain of +2.1 over the finetuned baseline and +4.3 over the pretrained model. 
The gains concentrate on \emph{Among} and \emph{Around}, where camera poses are object-centric and resemble the target-pose distribution produced by our view sampling (Sec.~\ref{sec:semantic_rendering}). In contrast, \emph{Rotation} consists of orthogonal views from an in-place rotating camera, which rarely occur among our sampled training views, and all finetuned variants fall below the pretrained model.

\clearpage
%%%%%%%%%%%%%%%%%%%%%%%%%%%%%%%%%%%%%%%%%%%%%%%%%%%%%%%%%%%%

\end{document}